\documentstyle[11pt,colacl]{article}

\title{Utilizing the World Wide Web as an Encyclopedia: \\ Extracting
Term Descriptions from Semi-Structured Texts}

\author{Atsushi Fujii \and Tetsuya Ishikawa \\
University of Library and Information Science \\
1-2 Kasuga, Tsukuba, 305-8550, JAPAN \\
\smallskip
{\normalsize\tt fujii@ulis.ac.jp}}

\input{psfig.tex}

\begin{document}

\maketitle\thispagestyle{empty}

\begin{abstract}
  In this paper, we propose a method to extract descriptions of
  technical terms from Web pages in order to utilize the World Wide
  Web as an encyclopedia. We use linguistic patterns and HTML text
  structures to extract text fragments containing term descriptions.
  We also use a language model to discard extraneous descriptions, and
  a clustering method to summarize resultant descriptions.  We show
  the effectiveness of our method by way of experiments.
\end{abstract}

\section{Introduction}
\label{sec:introduction}

Reflecting the growth in utilization of machine readable texts,
extraction and acquisition of linguistic knowledge from large corpora
has been one of the major topics within the natural language
processing (NLP) community. A sample of linguistic knowledge targeted
in past research includes grammars~\cite{kupiec:aaai-slnlp-ws-92},
word classes~\cite{hatzivassiloglou:acl-93} and bilingual
lexicons~\cite{smadja:cl-96}. While human experts find it difficult to
produce exhaustive and consistent linguistic knowledge, automatic
methods can help alleviate problems associated with manual
construction.

Term descriptions, which are usually carefully organized in
encyclopedias, are valuable linguistic knowledge, but have seldom been
targeted in past NLP literature.  As with other types of linguistic
knowledge relying on human introspection and supervision, constructing
encyclopedias is quite expensive. Additionally, since existing
encyclopedias are usually revised every few years, in many cases users
find it difficult to obtain descriptions for newly created terms.

To cope with the above limitation of existing encyclopedias, it is
possible to use a search engine on the World Wide Web as a substitute,
expecting that certain Web pages will describe the submitted
keyword. However, since keyword-based search engines often retrieve a
surprisingly large number of Web pages, it is time-consuming to
identify pages that satisfy the users' information needs.

In view of this problem, we propose a method to automatically extract
term descriptions from Web pages and summarize them. In this paper, we
generally use ``Web pages'' to refer to those pages containing textual
contents, excluding those with only image/audio information. Besides
this, we specifically target descriptions for technical terms, and
thus ``terms'' generally refer to technical terms.

In brief, our method extracts fragments of Web pages, based on
patterns (or templates) typically used to describe terms.  Web pages
are in a sense semi-structured data, because HTML (Hyper Text Markup
Language) tags provide the textual information contained in a page
with a certain structure. Thus, our method relies on both linguistic
and structural description patterns.

We used several NLP techniques to semi-automatically produce
linguistic patterns.  We call this approach ``NLP-based method.'' We
also produced several heuristics associated with the use of HTML tags,
which we call ``HTML-based method.'' While the former method is
language-dependent, and currently applied only to Japanese, the latter
method is theoretically language-independent.

Our research can be classified from several different perspectives. As
explained in the beginning of this section, our research can be seen
as linguistic knowledge extraction. Specifically, our research is
related to Web mining methods~\cite{nie:sigir-99,resnik:acl-99}.

From an information retrieval point of view, our research can be seen
as constructing domain-specific (or task-oriented) Web search engines
and software agents~\cite{etzioni:ai-magazine-97,mccallum:ijcai-99}.

\section{Overview}
\label{sec:overview}

Our objective is to collect encyclopedic knowledge from the Web, for
which we designed a system involving two processes. As with existing
Web search systems, in the background process our system periodically
updates a database consisting of term descriptions (a description
database), while users can browse term descriptions anytime in the
foreground process.

\begin{figure}[t]
  \begin{center}
    \leavevmode
    \psfig{file=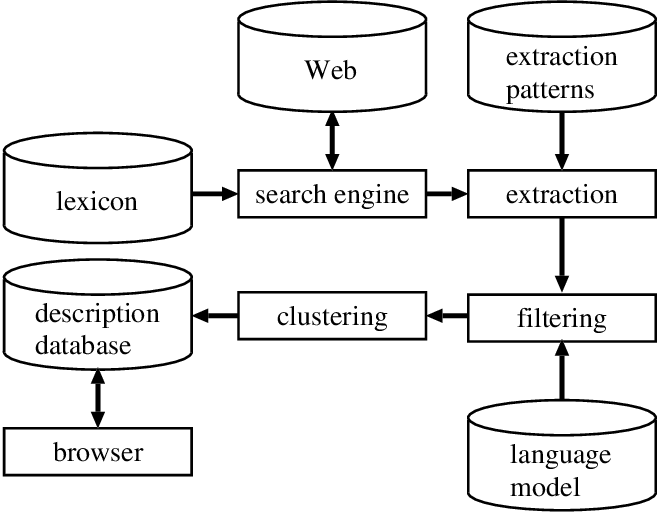,height=1.8in}
  \end{center}
  \caption{The control flow of our extraction system.}
  \label{fig:background}
\end{figure}

In the background process, depicted as in Figure~\ref{fig:background},
a search engine searches the Web for pages containing terms listed in
a lexicon.

Then, fragments (such as paragraphs) of retrieved Web pages are
extracted based on linguistic and structural description
patterns. Note that as a preprocessing for the extraction process, we
discard newline codes, redundant white spaces, and HTML tags that our
extraction method does not use, in order to standardize the layout of
Web pages.

However, in some cases the extraction process is unsuccessful, and
thus extracted fragments are not linguistically understandable.  In
addition, Web pages contain some non-linguistic information, such as
special characters (symbols) and e-mail addresses for contact, along
with linguistic information. Consequently, those noises decrease
extraction accuracy.

In view of this problem, we perform a filtering to enhance the
extraction accuracy. In practice, we use a language model to measure
the extent to which a given extracted fragment can be linguistic, and
index only fragments judged as linguistic into the description
database.

At the same time, the URLs of Web pages from which descriptions were
extracted are also indexed in the database, so that users can browse
the full content, in the case where descriptions extracted are not
satisfactory.

In the case where a number of descriptions are extracted for a single
term, the resultant description set is redundant, because it contains
a number of similar descriptions. Thus, it is preferable to summarize
descriptions, rather than to present all the descriptions as a list.

For this purpose, we use a clustering method to divide descriptions
for a single term into a certain number of clusters, and present only
descriptions that are representative for each cluster. As a result, it
is expected that descriptions resembling one another will be in the
same cluster, and that each cluster corresponds to different
viewpoints and word senses.

Possible sources of the lexicon include existing machine readable
terminology dictionaries, which often list terms, but lack
descriptions. However, since new terms unlisted in existing
dictionaries also have to be considered, newspaper articles and
magazines distributed via the Web can be possible sources. In other
words, a morphological analysis is performed periodically (e.g.,
weekly) to identify word tokens from those resources, in order to
enhance the lexicon. However, this is not the central issue in this
paper.

In the foreground process, given an input term, a browser presents one
or more descriptions to a user. In the case where the database does
not index descriptions for the given term, term descriptions are
dynamically extracted as in the background process. The background
process is optional, and thus term descriptions can always be obtained
dynamically. However, this potentially decreases the time efficiency
for a real-time response.

\begin{figure}[htbp]
  \begin{center}
    \leavevmode
    \psfig{file=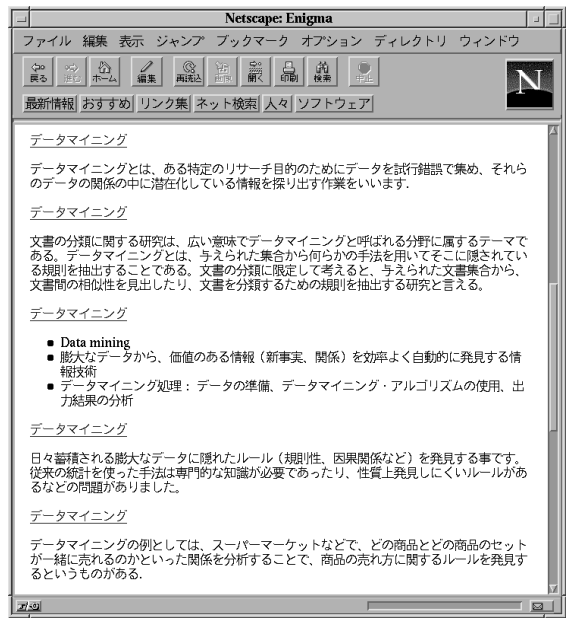,height=3.2in}
  \end{center}
  \caption{Example Japanese descriptions for ``{\it
  deeta-mainingu\/}~(data mining).''}
  \label{fig:enigma}
\end{figure}

Figure~\ref{fig:enigma} shows a Web browser, in which our prototype
page presents several Japanese descriptions extracted for the word
``{\it deeta-mainingu\/}~(data mining).'' For example, an English
translation for the first description is as follows:
\begin{quote}
  data mining is a process that collects data for a certain task, and
  retrieves relations latent in the data.
\end{quote}

In Figure~\ref{fig:enigma}, each description uses various expressions,
but describes the same content: data mining is a process which
discovers rules latent in given databases. It is expected that users
can understand what data mining is, by browsing some of those
descriptions. In addition, each headword (``{\it deeta-mainingu\/}''
in this case) positioned above each description is linked to the Web
page from which the description was extracted.

In the following sections, we first elaborate on the NLP/HTML-based
extraction methods in Section~\ref{sec:extraction}. We then elaborate
on noise reduction and clustering methods in Sections~\ref{sec:n-gram}
and \ref{sec:clustering}, respectively. Finally, in
Section~\ref{sec:experimentation} we investigate the effectiveness of
our extraction method by way of experiments.

\section{Extracting Term Descriptions}
\label{sec:extraction}

\subsection{NLP-based Extraction Method}
\label{subsec:nlp-based}

The crucial content for the NLP-based extraction method is the way to
produce linguistic patterns that can be used to describe technical
terms. However, human introspection is a difficult method to
exhaustively enumerate possible description patterns.

Thus, we used NLP techniques to semi-automatically collect description
patterns from machine readable encyclopedias, because they usually
contain a significantly large number of descriptions for existing
terms.  In practice, we used the Japanese CD-ROM World
Encyclopedia~\cite{heibonsha:98}, which includes approximately 80,000
entries related to various fields.

Before collecting description patterns, through a preliminary study on
the encyclopedia we used, we found that term descriptions frequently
contain salient patterns consisting of two Japanese ``{\it
bunsetsu\/}'' phrases. The following sentence, which describes the
term ``X,'' contains a typical {\it bunsetsu\/} combination, that is,
``X~{\it toha\/}'' and ``{\it de-aru\/}'':
\begin{list}{}{}
\item X {\it toha\/} Y {\it de-aru\/}~~~(X is Y).\footnote{Although
  ``{\it de-aru\/}'' itself is not a {\it bunsetsu\/} phrase, we use
  {\it bunsetsu\/} phrases to refer to combinations of several words.}
\end{list}
In other words, we collected description patterns, based on the
co-occurrence of two {\it bunsetsu\/} phrases, as in the following
method.

First, we collected entries associated with technical terms listed in
the World Encyclopedia, and replaced headwords with a variable ``X.''
Note that the World Encyclopedia describes various types of words,
including technical terms, historical people and places, and thus
description patterns vary depending on the word type. For example,
entries for historical people usually contain when/where the people
were born and their major contributions to the society.

However, for the purposes of our extraction, it is desirable to use
entries solely associated with technical terms. We then consulted the
EDR machine readable technical terminology dictionary, which contains
approximately 120,000 terms related to the information processing
field~\cite{edr-eng:95}, and obtained 2,259 entries associated with
terms listed in the EDR dictionary.

Second, we used the ChaSen morphological
analyzer~\cite{matsumoto:chasen-97}, which has commonly been used for
much Japanese NLP research, to segment collected entries into words,
and assign them parts-of-speech. We also developed simple heuristics
to produce {\it bunsetsu\/} phrases based on the part-of-speech
information.

Finally, we collected combinations of two {\it bunsetsu\/} phrases,
and sorted them according to their co-occurrence frequency, in
descending order. However, since the resultant {\it bunsetsu\/}
co-occurrences (even with higher rankings) are extraneous, we
supervised (verified, corrected or discarded) the top 100 candidates,
and produced 20 description patterns. Figure~\ref{fig:patterns} shows
a fragment of the resultant patterns and their English glosses. In
this figure, ``X'' and ``Y'' denote variables to which technical terms
and sentence fragments can be unified, respectively.

\begin{figure}[htbp]
  \def\baselinestretch{1}
  \begin{center}
    \leavevmode
    \small
    \begin{tabular}[t]{ll} \hline\hline
      {\hfill\centering Japanese\hfill}
      & {\hfill\centering English Gloss\hfill} \\ \hline
      X {\it toha\/} Y {\it dearu}. & X is Y. \\
      X {\it ha\/} Y {\it dearu}. & X is Y. \\
      Y {\it wo\/} X {\it to-iu}. & Y is called X. \\
      X {\it wo\/} Y {\it to-sadameru}. & X is defined as Y. \\
      Y {\it wo\/} X {\it to-yobu}. & Y is called X. \\
      \hline
    \end{tabular}
  \end{center}
  \caption{A fragment of linguistic description patterns we produced.}
  \label{fig:patterns}
\end{figure}

Here, we are in a position to extract sentences that match with
description patterns, from Web pages retrieved by the search engine
(see Figure~\ref{fig:background}). In this process, we do not conduct
morphological analysis on Web pages, because of computational
cost. Instead, we first segment textual contents in Web pages into
sentences, based on the Japanese punctuation system, and use a surface
pattern matching based on regular expressions.

However, in most cases term descriptions consist of more than one
sentence. This is especially salient in the case where anaphoric
expressions and itemization are used. Thus, it is desirable to extract
a larger fragment containing sentences that match with description
patterns.

In view of this problem, we first use linguistic description patterns
to briefly identify a zone, and sequentially search the following
fragments relying partially on HTML tags, until a certain fragment is
extracted:\footnote{Although we use HTML tags to identify appropriate
text fragments, we call the method described in this section NLP-based
method, in a comparison with the method in
Section~\ref{subsec:html-based} that relies solely on HTML tags.}
\begin{enumerate}
  \def\labelenumi{(\theenumi)}
\item paragraph tagged with \verb$<P>...</P>$ (or
  \verb$<P>...<P>$ in the case where \verb$</P>$ is missing),
\item itemization tagged with \verb$<UL>...</UL>$,
\item $N$ sentences identified with the Japanese punctuation system,
  where the sentence that matched with a description pattern is
  positioned as near center as possible, where we empirically set
  \mbox{$N=3$}.
\end{enumerate}

\subsection{HTML-based Extraction Method}
\label{subsec:html-based}

Through a preliminary study on existing Web pages, we identified two
typical usages of HTML tags associated with describing technical
terms.

In the first usage, a term in question is highlighted as a heading by
way of \verb$<H>...</H>$, \verb$<B>...</B>$ or \verb$<DT>$ tag, and
followed by its description in a short fragment. In the second usage,
terms that are potentially unfamiliar to readers are tagged with the
anchor \verb$<A>$ tag, providing hyperlinks to other pages (or a
different position within the same page) where they are described.

The crucial factor here is to determine which fragment in the page is
extracted as a description. For this purpose, we use the same rules
described in Section~\ref{subsec:nlp-based}. However, unlike the
NLP-based method, in the HTML-based method we extract the fragment
that {\em follows\/} the heading and the position linked from the
anchor. However, in the case where a term in question is tagged with
\verb$<DT>$, we extract the following fragment tagged with
\verb$<DD>$. Note that \verb$<DT>$ and \verb$<DD>$ are inherently
provided to describe terms.

The HTML-based method is expected to extract term descriptions that
cannot be extracted by the NLP-based method, and vice versa. In fact,
in Figure~\ref{fig:enigma} the third and fourth descriptions were
extracted with the HTML-based method, while the rest were extracted
with the NLP-based method.

\section{Language Modeling for Filtering}
\label{sec:n-gram}

Given a set of Web page fragments extracted by the NLP/HTML-based
methods, we select fragments that are linguistically understandable,
and index them into the description database. For this purpose, we
perform a language modeling, so as to quantify the extent to which a
given text fragment is linguistically acceptable.

There are several alternative methods for language modeling. For
example, grammars are relatively strict language modeling methods.
However, we use a model based on $N$-gram, which is usually more
robust than that based on grammars. In other words, text fragments
with lower perplexity values are more linguistically acceptable.

In practice, we used the CMU-Cambridge
toolkit~\cite{clarkson:eurospeech-97}, and produced a trigram-based
language model from two years of \mbox{Mainichi Shimbun} Japanese
newspaper articles~\cite{mainichi:94-95}, which were automatically
segmented into words by the ChaSen morphological
analyzer~\cite{matsumoto:chasen-97}.

In the current implementation, we empirically select as the final
extraction results text fragments whose perplexity values are lower
than 1,000.

\section{Clustering Term Descriptions}
\label{sec:clustering}

For the purpose of clustering term descriptions extracted using
methods in Sections~\ref{sec:extraction} and \ref{sec:n-gram}, we use
the Hierarchical Bayesian Clustering (HBC)
method~\cite{iwayama:ijcai-95}, which has been used for clustering
news articles and constructing thesauri.

As with a number of hierarchical clustering methods, the HBC method
merges similar items (i.e., term descriptions in our case) in a
bottom-up manner, until all the items are merged into a single
cluster. That is, a certain number of clusters can be obtained by
splitting the resultant hierarchy at a certain level.

At the same time, the HBC method also determines the most
representative item (centroid) for each cluster. Then, we present only
those centroids to users.

The similarity between items is computed based on feature vectors that
characterize each item. In our case, vectors for each term description
consist of frequencies of content words (e.g., nouns and verbs
identified through a morphological analysis) appearing in the
description.

\section{Experimentation}
\label{sec:experimentation}

\subsection{Methodology}
\label{subsec:experiment_method}

We investigated the effectiveness of our extraction method from a
scientific point of view. However, unlike other research topics where
benchmark test collections are available to the public (e.g.,
information retrieval), there are two major problems for the purpose
of our experimentation, as follows:
\begin{itemize}
\item production of test terms for which descriptions are extracted,
\item judgement for descriptions extracted for those test terms.
\end{itemize}
For test terms, possible sources are those listed in existing
terminology dictionaries. However, since the judgement can be
considerably expensive for a large number of test terms, it is
preferable to selectively sample a small number of terms that
potentially reflect the interest in the real world.

In view of this problem, we used as test terms those contained in
queries in the NACSIS test collection~\cite{kando:sigir-99}, which
consists of 60 Japanese queries and approximately 330,000 abstracts
(in either a combination of English and Japanese or either of the
languages individually), collected from technical papers published by
65 Japanese associations for various fields.\footnote{\tt
{http://www.rd.nacsis.ac.jp/\~{}ntcadm/\\index-en.html}}

This collection was originally produced for the evaluation of
information retrieval systems, where each query is used to retrieve
technical abstracts. Thus, the title field of each query usually
contains one or more technical terms. Besides this, since each query
was produced based partially on existing technical abstracts, they
reflect the real world interest, to some extent.  As a result, we
extracted 53 test terms, as shown in Table~\ref{tab:result}. In this
table, we romanized Japanese terms, and inserted hyphens between each
morpheme for enhanced readability.

Note that unlike the case of information retrieval (e.g., a patent
retrieval), where every relevant document must be retrieved, in our
case even one description can potentially be sufficient. In other
words, in our experiments, more weight is attached to accuracy
(precision) than recall.

For the search engine in Figure~\ref{fig:background}, we used
``goo,''\footnote{{\tt http://www.goo.ne.jp/}} which is one of the
major Japanese Web search engines.  Then, for each extracted
description, one of the authors judged it correct or incorrect.

\subsection{Results}
\label{subsec:experiment_result}

Out of the 53 test terms extracted from the NACSIS collection, for 44
terms goo retrieved one or more Web pages.  Among those 44 test terms,
our method extracted at least one term description for 27 terms,
disregarding the judgement. Thus, the coverage (or applicability) of
our method was 61.4\%. In Table~\ref{tab:result}, the third column
denotes the number of Web pages identified by goo. However, goo
retrieves contents for only the top 1,000 pages.

Table~\ref{tab:result} also shows the number descriptions judged as
correct (the column ``\#C''), the total number of descriptions
extracted (the column ``\#T''), and the accuracy (the column ``A''),
for both cases with/without the trigram-based language model.

\begin{table*}[htbp]
  \def\baselinestretch{1}
  \begin{center}
    \caption{Extraction accuracy for the 27 test terms (\#C = the 
    number of correct descriptions, \#T = the total number of
    extracted descriptions, A = accuracy (\%)).}
    \medskip
    \leavevmode
    \footnotesize
    \tabcolsep=3pt
    \begin{tabular}{llrrrrrrr} \hline\hline
      & & & \multicolumn{3}{c}{w/o Trigram} & \multicolumn{3}{c}{w
      Trigram} \\ \cline{4-9}
      {\hfill\centering Japanese Term\hfill} &
      {\hfill\centering English Gloss\hfill} &
      {\hfill\centering \#Pages\hfill}
      & {\hfill\centering \#C\hfill} & {\hfill\centering \#T\hfill} &
      {\hfill\centering A\hfill} & {\hfill\centering \#C\hfill} &
      {\hfill\centering \#T\hfill} & {\hfill\centering A\hfill} \\
      \hline
      Zipf{\it -no-housoku\/} & Zipf's law & 15 & 1 & 1 & 100 & 1 & 1 &
      100 \\
      {\it akusesu-seigyo\/} & access control & 6,925 & 10 & 20 & 50.0
      & 10 & 20 & 50.0 \\
      {\it bunsho-gazou-rikai\/} & document image understanding & 43 &
      1 & 1 & 100 & 1 & 1 & 100 \\
      {\it chiteki-eejento\/} & intelligent agent & 323 & 3 & 5 & 60.0
      & 3 & 5 & 60.0 \\
      {\it deeta-mainingu\/} & data mining & 3,389 & 37 & 49 &
      75.5 & 30 & 40 & 75.0 \\
      {\it denshi-sukashi\/} & digital watermark & 2,124 & 29 & 32 &
      90.6 & 29 & 32 & 90.6 \\
      {\it denshi-toshokan\/} & digital library & 7,938 & 10 & 26 &
      38.5 & 8 & 17 & 47.1 \\
      {\it gazou-kensaku\/} & image retrieval & 1,694 & 1 & 4 & 25.0 &
      1 & 3 & 33.3 \\
      {\it guruupuwea\/} & groupware & 19,760 & 14 & 40 & 35.0 & 12 &
      21 & 57.1 \\
      {\it hikari-faibaa\/} & optical fiber & 10,078 & 17 & 25 & 68.0
      & 14 & 21 & 66.7 \\
      {\it ichi-keisoku\/} & position measurement & 735 & 0 & 3 & 0 &
      0 & 3 & 0 \\
      {\it identeki-arugorizumu\/} & genetic algorithm & 4,686 & 24 &
      31 & 77.4 & 22 & 28 & 78.6 \\
      {\it jinkou-chinou\/} & artificial intelligence & 18,190 & 10 &
      19 & 52.6 & 9 & 13 & 69.2 \\
      {\it jiritsu-idou-robotto\/} & autonomous mobile robot & 792 & 2
      & 2 & 100 & 2 & 2 & 100 \\
      {\it jisedai-intaanetto\/} & next generation Internet & 1,963 &
      6 & 10 & 60.0 & 6 & 10 & 60.0 \\
      {\it kiiwaado-jidou-chuushutsu\/} & keyword automatic extraction
      & 25 & 1 & 1 & 100 & 1 & 1 & 100 \\
      {\it kikai-hon'yaku\/} & machine translation & 3,141 & 1 & 10 &
      10.0 & 0 & 8 & 0 \\
      {\it korokeishon\/} & collocation & 547 & 7 & 16 & 43.8 &
      7 & 15 & 46.7 \\
      {\it koshou-shindan\/} & fault diagnosis & 1,682 & 2 & 5 & 40.0 &
      2 & 4 & 50.0 \\
      {\it maruchikyasuto\/} & multicast & 5,758 & 18 & 25 & 72.0 & 15
      & 22 & 68.2 \\
      {\it media-douki\/} & media synchronization & 46 & 1 & 1 & 100 & 1
      & 1 & 100 \\
      {\it nettowaaku-toporojii\/} & network topology & 438 & 1 & 4 &
      25.0 & 0 & 3 & 0 \\
      {\it nyuuraru-nettowaaku\/} & neural network & 9,537 & 37 & 47 &
      78.7 & 36 & 45 & 80.0 \\
      {\it ringu-gata-nettowaaku\/} & ring network & 44 & 0 & 1 & 0 &
      0 & 1 & 0 \\
      {\it shisourasu\/} & thesaurus & 3,399 & 21 & 23 &
      91.3 & 19 & 20 & 95.0 \\
      {\it souraa-kaa\/} & solar car & 3,698 & 12 & 21 &
      57.1 & 12 & 21 & 57.1 \\
      {\it teromea\/} & telomere & 873 & 26 & 36 & 72.2 &
      25 & 34 & 73.5 \\
      \hline
      {\hfill\centering total\hfill} & {\hfill\centering ---\hfill} &
      109,049 & 292 & 460 & 63.5 & 266 & 392 & 67.9 \\ \hline
    \end{tabular}
    \label{tab:result}
  \end{center}
\end{table*}

Table~\ref{tab:result} shows that the NLP/HTML-based methods extracted
appropriate term descriptions with a 63.5\% accuracy, and that the
trigram-based language model further improved the accuracy from 63.5\%
to 67.9\%. In other words, only two descriptions are sufficient for
users to understand a term in question. Reading a few descriptions is
not time-consuming, because they usually consist of short paragraphs.

We also investigated the effectiveness of clustering, where for each
test term, we clustered descriptions into three clusters (in the case
where there are less than four descriptions, individual descriptions
were regarded as different clusters), and only descriptions determined
as representative by the HBC method were presented as the final
result. We found that 66.7\% of descriptions presented were correct
ones.  In other words, users can obtain descriptions from different
viewpoints and word senses, maintaining the extraction accuracy
obtained above (i.e., 67.9\%).

However, we concede that we did not investigate whether or not each
cluster corresponds to different viewpoints in a rigorous manner.

For the polysemy problem, we investigated all the descriptions
extracted, and found that only ``{\it korokeishon\/}~(collocation)''
was associated with two word senses, that is, ``word collocations''
and ``position of machinery.''  Among the three representative
descriptions for ``{\it korokeishon\/}~(collocation),'' two
corresponded to the first sense, and one corresponded to the second
sense. To sum up, the HBC clustering method correctly identified
polysemy.

\section{Conclusion}
\label{sec:conclusion}

In this paper, we proposed a method to extract encyclopedic knowledge
from the World Wide Web.

For extracting fragments of Web pages containing term descriptions, we
used linguistic and HTML structural patterns typically used to describe
terms. Then, we used a language model to discard irrelevant
descriptions. We also used a clustering method to summarize extracted
descriptions based on different viewpoints and word senses.

We evaluated our method by way of experiments, and found that the
accuracy of our extraction method was practical, that is, a user can
understand a term in question, by browsing two descriptions, on
average. We also found that the language model and the clustering
method further enhanced our framework.

Future work will include experiments using a larger number of test
terms, and application of extracted descriptions to other NLP
research.

\section*{Acknowledgments}

The authors would like to thank Hitachi Digital Heibonsha, Inc. for
their support with the CD-ROM World Encyclopedia, Makoto Iwayama and
Takenobu Tokunaga for their support with the HBC clustering software,
and Noriko Kando (National Institute of Informatics, Japan) for her
support with the NACSIS collection.

\bibliographystyle{acl}

\end{document}